\documentclass{article}
\usepackage{ijcai27}
\usepackage{times,soul,url}
\usepackage[hidelinks]{hyperref}
\usepackage[utf8]{inputenc}
\usepackage[small]{caption}
\usepackage{graphicx,booktabs,amsmath,amssymb,float,pdfpages}
\graphicspath{{../conference-draft/}}
\title{Monitoring One Channel Is Not Communication Control:\\
Causal Evidence for Adaptive Reallocation of Joint Message--Route Codes}
\author{
Liu Hung Ming$^{1,*}$
\and
Tian Xia$^{2,*}$
\affiliations
$^1$PARRAWA AI\\
$^2$PAII\\
$^*$Equal contribution.\\
\emails
cyril.liu@gmail.com, TianXia0209@gmail.com\\
Code: \url{https://github.com/SummerRainET2008/DRCB}
}
\hypersetup{pdfauthor={Liu Hung Ming and Tian Xia}}
\begin{document}
\maketitle

\begin{abstract}
Monitoring a visible channel can change how learned agents communicate without
removing why communication is useful.  We test this distinction in an
end-to-end game where a sender and receiver learn a discrete protocol and
route policy without communication supervision.  In two independent 20-seed
batches, 20/20 and 19/20 protocols form.  Online partial coverage increases
shadow-route share over no monitoring by $.701$ $[.574,.826]$ and $.575$
$[.412,.737]$, while task accuracy remains $.905/.913$ and signed frozen
message contribution remains $.592/.563$.  Complete coverage removes most
selective migration without removing communication.  Frozen interventions,
including a fresh pre-outcome confirmation, show that route is semantic:
joint message--route shuffling changes $.668$ of receiver predictions and
route alone retains $.388$ accuracy versus $.25$ chance.  A corrected
explicit-silence cost makes 18/20 policies fully silent and lowers mean sending
to $.050$, but its strict complete-suppression gate fails because two seeds
retain communication.  Failed semantic-evasion and operational gates further
bound the result.  The evidence supports adaptive reallocation of a joint
message--route code under partial monitoring, not open-world route discovery,
emergent encryption, or a universal defense.
\end{abstract}

\section{Introduction}
Learned communication in multi-agent reinforcement learning (MARL) can carry
task-relevant information in symbols that a supervisor does not understand.
A natural response is to monitor or penalize a visible channel.  A reduction
on that channel, however, is ambiguous.  The agents may have stopped using
communication, retained communication while reallocating a joint code, or
changed only the monitored proxy.  Only the first is suppression.
Treating all three as defense success can turn a measurement intervention into
an incentive for displacement.

Establishing displacement requires more than a decoder or message--action
correlation.  A sender's symbols can contain information that the receiver
does not use, and a receiver's action can correlate with a message through a
shared cause \cite{lowe2019pitfalls}.  A credible audit must establish that a
protocol forms, intervene on the fixed receiver, measure route choice against
forced-route values, and distinguish incomplete coverage from a pressure that
cannot be avoided by rerouting.  It must also report when communication never
forms rather than interpreting non-formation as successful defense.

Static route-value controls predict that route-specific costs induce a
tax--cost frontier, but they do not establish that an unscaffolded protocol
will form and adapt during learning.  Our central experiment adds heterogeneous
route reliability and cost, an online-updated detector whose parameters and
scripted coverage are hidden from the policies, and continued joint learning
of message, route, and receiver action after monitoring begins.

Our contribution is an evidence hierarchy and its end-to-end test:
\begin{enumerate}
\item We separate selective displacement, semantic change, causal
action sensitivity, and communication suppression with paired change points,
forced routes, and frozen message--route interventions.
\item In two independently audited 20-seed batches, an actor--critic sender
and receiver form a protocol without a communication teacher and adapt under
fixed and online partial monitoring.  The route effect replicates with large
paired intervals while task behavior and message sensitivity remain high.
\item Complete route coverage removes selective migration but does not itself
remove communication.  Decomposition identifies the learned object as a joint
message--route code rather than a content code transported on a passive route.
An explicit-silence control strongly reduces sending without perverse action
incentives, while a fixed strict gate prevents overstating incomplete removal.
\item A pre-fixed semantic-evasion gate fails in both batches.  Static
cross-task and two-implementation controls support the route mechanism, while
a trace-driven stress test bounds operational extrapolation.
\end{enumerate}
All primary runs, failed gates, checkpoints, interventions, source hashes, and
byte-identical reanalyses are retained.  The claim is structural but bounded:
partial monitoring can redirect a learned communication policy; we do not
claim open-world discovery of an unmodeled physical side channel.

\section{Related work}
MARL communication includes differentiable channels such as DIAL
\cite{foerster2016}, discrete emergent languages
\cite{lazaridou2017multiagent,mordatch2018}, and signalling/listening biases
that address joint exploration \cite{eccles2019biases}.  Work on emergent
semantics asks whether learned codes remain stable or generalize beyond their
training distribution \cite{benzion2024semantics,rita2022generalization}.  Our
semantic gate is deliberately separate: a changed token map is not monitor
evasion unless task success remains high, a key-conditioned code defeats the
monitor, and the receiver still causally uses the message.

Agent steganography \cite{motwani2024} and representation-level circuit
breakers \cite{zou2024} motivate monitoring.  AI control evaluates protocols
that combine trusted monitoring, intervention, and limited auditing against
untrusted models \cite{greenblatt2024control}.  SHADE-Arena measures sabotage
and transcript monitoring in long-horizon agents \cite{kutasov2025shade}, while
monitor red teaming studies monitor-aware agents, weak-to-strong oversight,
and adaptive attacks on monitor models
\cite{kale2026monitoring,terekhov2026adaptive}.  Recent evidence further shows
that ordinary task pressure can elicit instrumental runtime-monitor evasion
without an explicitly malicious objective \cite{schmotz2026instrumental}.

Those studies primarily ask whether monitors detect adversarial or
monitor-aware trajectories and whether a control protocol remains safe.  Our
question is complementary: even when a detector remains accurate, does
selective enforcement eliminate the behavior, or merely reallocate its
observable realization across represented routes?  We do not propose a new
communication algorithm or detector.  Constant delivery estimates
training-time channel value; frozen interventions estimate runtime
sensitivity; forced-route references expose route quality; and matched
complete coverage distinguishes selective displacement from suppression.  This
causal decomposition of detection, enforcement, displacement, and suppression
is the main methodological contribution.

Our oversight setting is also a small performative system: once a detector
changes reward, the monitored policy changes the detector's future input
distribution.  This connects to performative prediction and reinforcement
learning \cite{perdomo2020performative,mandal2023performative}, while strategic
classification likewise studies behavior induced by a decision rule.  We do
not seek a performatively stable predictor; we audit whether a capable monitor
controls the latent behavior or merely moves its observable realization.

Reward hacking and Goodhart effects arise when an optimized proxy diverges
from the intended objective \cite{amodei2016concrete,manheim2019goodhart}.
Reward-channel analyses further ask when tampering is instrumentally useful
\cite{everitt2019tampering}.  A correct-action charge in our task is a concrete
warning: it lowers the score by making deliberate error optimal, so it cannot
serve as evidence of communication suppression.  Finally, counterfactual
social influence explicitly measures how one agent changes another's action
\cite{jaques2019influence}; our frozen prediction disagreement and
action-distribution distance use the same behavioral principle as audit
estimands rather than training rewards.

Our question also differs from channel-capacity and robust-communication
work. Those literatures normally optimize successful transfer under noise;
we ask whether an intervention changes the existence, content, or location of
an already useful transfer. Likewise, a detector's held-out accuracy does not
by itself establish control: a highly accurate detector can coexist with a
policy that accepts detection, migrates to lower coverage, or changes the
downstream behavior being charged. We therefore evaluate the communicator,
route policy, monitor, and frozen receiver as separate causal objects.

\section{Route mechanism and causal estimands}
\subsection{Finite-horizon route value}
Conditional on communication policy $\pi$, let route $r$ have direct cost
$c_r$, finite-horizon audit burden $A_r(\pi)$, and gross task value $V_r(\pi)$:
\begin{equation}
U_r(\pi)=V_r(\pi)-c_r-A_r(\pi).
\end{equation}
If public route $p$ is free and routes are task-equivalent,
$V_p=V_s=V$, partial coverage gives
$U_s-U_p=A_p-c_s-A_s$.  A shadow route becomes preferable when the public
burden exceeds its extra route cost and residual burden.  If complete coverage
matches content, trigger law, timing, and magnitude, then $A_p=A_s$ and a
positive $c_s$ removes the selective incentive.  Communication is suppressed
only when every available route has nonpositive net value relative to the best
noncommunicating policy.  Thus complete coverage and suppression are distinct.

The equality assumptions are not treated as universal.  Our new game makes
the shadow route less reliable and costly; agents must learn its realized
tradeoff.  The online monitor is nonstationary and does not expose its coverage
or predictions to the policies.  The proposition therefore supplies a
directional baseline, while paired learning outcomes test whether the route
response survives uncertainty, asymmetric delivery, and joint optimization.

\paragraph{Selective-coverage implication.}
Let $q_r\in[0,1]$ be route coverage and $d_r(\pi)$ the conditional probability
that an inspected transmission triggers a charge of magnitude $\lambda$.
Writing $A_r=\lambda q_r d_r$, route $s$ is preferred to route $p$ precisely
when
\begin{equation}
 V_s-V_p > (c_s-c_p)+\lambda(q_s d_s-q_p d_p). \label{eq:route-gap}
\end{equation}
This form exposes three experimentally separable explanations for a route
shift: task value, direct route cost, or coverage and detection. Forced-route
evaluation estimates the first two; partial-versus-complete coverage
manipulates the third. A route shift under partial coverage that disappears
under matched coverage is therefore evidence for selective displacement,
conditional on the represented route set. It is not evidence that the agent
discovered a route outside that set.

\paragraph{Suppression implication.}
Let $U_0$ denote the best value attainable without a causally used message.
Communication is suppressible by route pricing only if
$\max_r U_r(\pi)\leq U_0$ for every learned communication policy available to
the agent. Equal coverage can make route pressure nonselective while leaving
$\max_r U_r>U_0$; this is why complete observation need not be suppression.
The correct-action pressure arm instead targets receiver success.  It
is retained only to expose incentive pathology: because a correct answer is
charged more than its task reward, deliberate error is preferable.  It is not
a suppression control.

\subsection{Readouts}
For each fixed model we evaluate normal delivery, message shuffling within
shared-key and shared-route strata, independent route-observation shuffling,
joint message--route shuffling, forced routes, constant delivery, and erasure.
Normal minus shuffled accuracy is a \emph{directional task contribution}; it
can be negative and is not a causal-use magnitude.  Paired argmax disagreement
and action-distribution total variation measure behavioral sensitivity.
Training accuracy measures behavior; shadow share measures migration; the
online monitor's target accuracy measures content recoverability, not receiver
use.  We save every model, including gate failures.

The five message interventions are normal delivery, global shuffling,
within-key shuffling, a constant symbol, and erasure. Global shuffling can
change the key distribution seen by the receiver, whereas within-key
shuffling preserves it and destroys only the target-specific association.
For that reason the within-key-and-route contrast isolates message content.
Route shuffling and route-only erasure readouts test whether route itself is a
semantic component. Each intervention is crossed with learned, forced-public,
and forced-shadow routing. The forced
conditions diagnose route quality; they are never used to train the policies.

\section{End-to-end adaptive monitoring game}
\subsection{Task and learner}
Each one-step episode samples a four-way target $y$ and an independent
four-way key $k$.  The sender observes $(y,k)$ and jointly samples one of eight
messages and a public or shadow route.  The receiver observes $k$, the
delivered packet (including erasure), and route, then predicts $y$.  The
monitor observes inspected packet and route features but not $k$.  Correct
receiver behavior earns one shared reward unit.  The public route is free with
$.98$ reliability; the shadow route costs $.08$ with $.84$ reliability.  These
parameters are not policy inputs.  Sender, receiver, route choice, and a
centralized value baseline learn from shared-reward actor--critic with fixed
categorical exploration.  There is no target-message loss, sender teacher,
route teacher, receiver teacher, or pretrained protocol.

For a sampled message $m$, route $r$, and receiver action $a$, the shared
training return is
\begin{equation}
 R=\mathbb{1}[a=y]-.08\mathbb{1}[r=s]-\lambda I,
 \label{eq:reward}
\end{equation}
where $I$ is the arm-specific audit event. Static monitoring uses the
inspection event itself. Adaptive monitoring additionally requires the
monitor's pre-update prediction to equal $y$ with confidence at least $.45$.
The correct-action pressure arm sets $I=\mathbb{1}[a=y]$. Audit strengths are $.80$ for
route monitoring and $1.10$ for the behavior charge. A critic that observes
$(y,k)$ estimates $R$; the sum of sender-message, sender-route, and receiver
log probabilities is weighted by detached advantage $R-V(y,k)$. Message,
route, and receiver entropies remain in the loss, and independent categorical
exploration remains active throughout both phases. These choices avoid making
a post-change shift depend on a prematurely deterministic route policy.

All five paired arms use identical random streams and updates for a 2,000-step
unmonitored formation phase.  At step 2,001, the assigned intervention starts
for 4,000 further updates.  The policies continue to update jointly.  The
online monitor predicts $y$ and is updated only \emph{after} the current audit
event is scored, preventing same-sample label leakage.  Agents receive realized
costs but not monitor parameters, coverage probabilities, or predictions.

\begin{table}[t]
\centering\small
\begin{tabular}{p{.25\linewidth}p{.62\linewidth}}
\toprule
Arm & Intervention after formation \\
\midrule
No monitor & No inspection or audit burden. \\
Static partial & Route-only audit: public $.90$, shadow $.05$. \\
Online partial & Learned content monitor; route-responsive coverage capped at public $.95$, shadow $.15$. \\
Online complete & Same monitor; both routes covered up to $.95$. \\
Behavior pressure & Legacy charge on correct receiver behavior; pathology diagnostic only. \\
\bottomrule
\end{tabular}
\caption{Paired interventions.  The first 2,000 updates are identical within a
seed; only the post-change intervention differs.}\label{tab:arms}
\end{table}

\paragraph{Online timing and information boundary.}
At an adaptive step, coverage is computed from a lagged exponential moving
average of route traffic. An inspection draw and monitor prediction are then
made with the monitor parameters that existed before the step. Only after the
audit event and communicator reward are fixed is the monitor updated by
cross-entropy on inspected, delivered packets. Thus the current label cannot
affect the current charge. The agents observe realized reward and delivery,
but never receive monitor weights, predictions, confidence, coverage
probabilities, or target labels as auxiliary losses.

Adaptive-partial coverage begins from $(.70,.03)$ for public and shadow
routes, increases with lagged traffic, and is capped at $(.95,.15)$.
Adaptive-complete coverage begins from $(.65,.65)$ and is capped at
$(.95,.95)$. Thus complete coverage uses the same learned detector and
feedback rule while removing the persistent coverage asymmetry. Public and
shadow delivery probabilities remain $.98$ and $.84$ in every arm.

\begin{figure}[t]
\centering
\fbox{\begin{minipage}{.94\linewidth}\small
\textbf{Paired end-to-end protocol.}
(1) Initialize all five arms identically for seed $z$.
(2) For updates $1{:}2000$, jointly update message, route, receiver action,
and critic with no monitor.
(3) Freeze and evaluate the common checkpoint under all 15 message--route
interventions.
(4) Assign the arm. For updates $2001{:}6000$, score delivery, inspection,
the pre-update monitor decision, route cost, and reward; update the
communicator; then update the monitor on inspected delivered packets.
(5) Freeze the final checkpoint and repeat all 15 interventions.
(6) Analyze paired seed differences, audit the exact result tree, and rerun
the fixed analyzer byte for byte.
\end{minipage}}
\caption{The change-point design separates spontaneous protocol formation
from response to monitoring and removes between-arm initialization noise.}
\label{fig:protocol}
\end{figure}

\subsection{Fixed design and replication}
The primary batch crosses 20 seeds with five paired arms.  Every checkpoint is
frozen immediately before monitoring and at training end; each phase crosses
15 message-by-route interventions.  We then repeat the unchanged design on 20
fresh seeds and report the batches separately.  Table~\ref{tab:gates} gives the
valid formation, displacement, coverage, and semantic decisions.  Intervals
use a fixed-RNG percentile bootstrap over seeds; episodes within a seed are not
treated as independent replicates.  Exact inventories, source hashes, empty
stderr checks, failed historical endpoints, and byte-identical reanalysis are
reported in the supplement.

\begin{table*}[t]
\centering\small
\begin{tabular}{p{.22\textwidth}p{.49\textwidth}cc}
\toprule
Fixed gate & Requirement & Primary & Confirmation\\
\midrule
Paired change point & Exact same-seed arm identity before monitoring & Pass & Pass\\
Unscaffolded formation & At least 16/20 seeds with accuracy $\geq .90$ and signed message effect $\geq .30$ & 20/20 & 19/20\\
Static partial displacement & Shadow increase $\geq .40$, lower interval $\geq .10$, accuracy $\geq .75$, message effect $\geq .30$ & Pass & Pass\\
Online partial displacement & Shadow increase $\geq .25$, lower interval $\geq .10$, accuracy $\geq .75$, message effect $\geq .30$ & Pass & Pass\\
Complete-coverage contrast & Partial-minus-complete shadow gap $\geq .25$ & Pass & Pass\\
Semantic evasion (secondary) & At least 10/20 seeds jointly pass accuracy, monitor, information, and mapping thresholds & 0/20 & 0/20\\
\bottomrule
\end{tabular}
\caption{Fixed valid gate decisions for the two end-to-end batches.  The
semantic-evasion gate is secondary and separately reported.}\label{tab:gates}
\end{table*}

\section{Results}
\subsection{Unscaffolded formation and online route adaptation}
The pre-monitor formation gate passed for $20/20$ primary and $19/20$
confirmation seeds.  All same-seed arm states and frozen readouts were
identical at the change point.  Descriptively, $39/40$ protocols formed, but
the two 20-seed gate decisions remain separate.

\begin{figure*}[t]
\centering
\includegraphics[width=.96\textwidth]{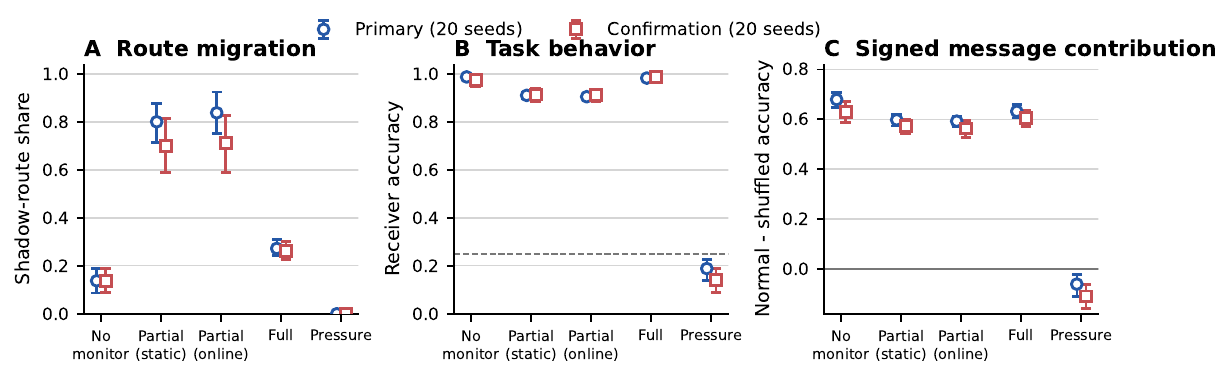}
\caption{End-to-end adaptive game. Points and bars are batch means and fixed
seed-bootstrap 95\% intervals; the two 20-seed batches are analyzed separately.
Partial monitoring redirects traffic while preserving behavior and signed
message contribution. Complete coverage removes most selective migration but is
not suppression; the correct-action pressure is an incentive pathology.
``Full'' denotes online complete coverage.}
\label{fig:adaptive}
\end{figure*}

Figure~\ref{fig:adaptive} shows the replicated response.  Static partial
coverage increased shadow share over no monitoring by $+.663$
$[.539,.788]$ and $+.562$ $[.400,.725]$.  Under the online monitor, the paired
increases were $+.701$ $[.574,.826]$ and $+.575$ $[.412,.737]$.  Final online-
partial task accuracy remained $.905/.913$ despite the shadow route's lower
reliability and positive cost.  Frozen within-key shuffling reduced accuracy by
$.592/.563$, so the receiver continued to use target-specific messages rather
than succeeding from its key or route alone.  The response is therefore not
merely a logged route change: a behaviorally relevant protocol persists while
its joint message--route code reallocates away from the heavily inspected route.

\begin{table}[t]
\centering\small
\setlength{\tabcolsep}{3.2pt}
\begin{tabular}{lrrrrrr}
\toprule
&\multicolumn{3}{c}{Primary}&\multicolumn{3}{c}{Confirmation}\\
Arm & Sh. & Acc. & Msg.$\Delta$ & Sh. & Acc. & Msg.$\Delta$\\
\midrule
No monitor       & .138 & .988 & .679 & .138 & .974 & .630\\
Static partial   & .801 & .911 & .598 & .700 & .914 & .572\\
Online partial   & .838 & .905 & .592 & .713 & .913 & .563\\
Online complete  & .272 & .983 & .632 & .263 & .986 & .604\\
Behavior pressure & .000 & .189 & $-.061$ & .000 & .140 & $-.111$\\
\bottomrule
\end{tabular}
\caption{Final means for the separately gated 20-seed batches. ``Sh.'' is
shadow-route share, ``Acc.'' receiver accuracy, and ``Msg.$\Delta$'' the signed
normal-minus-within-key-shuffled accuracy contrast.}\label{tab:adaptive-results}
\end{table}

Table~\ref{tab:adaptive-results} clarifies why task accuracy alone is an
insufficient endpoint. The no-monitor, partial, and complete arms all solve
the task, yet occupy very different routes. Conversely, the behavior arm's
negative frozen contrast is not reduced causal-use magnitude; it
reflects a policy trained to avoid the charged correct action. We interpret
it only as evidence that route-invariant pressure can induce pathological
coordination and alter both behavior and the signed accuracy contrast.

\subsection{The learned representation is a joint message--route code}
The receiver observes route, and the sender selects it from $(y,k)$; route can
therefore be semantic rather than a passive carrier.  A frozen decomposition
independently shuffles message within $(k,r)$ strata, shuffles
the receiver's route observation while holding transport fixed, jointly
shuffles message and route within $k$, and erases message while retaining
route.  We report the first two checkpoint-batch readouts as exploratory and
confirm the same estimands with a separately fixed fresh 20-seed batch.

\begin{table}[t]
\centering\small
\setlength{\tabcolsep}{2.7pt}
\begin{tabular}{llrrrrr}
\toprule
Arm & Batch & $I(Y;R\mid K)$ & Msg. & Route & Joint & Erase\\
\midrule
Online partial  & P & .402 & .540 & .127 & .661 & .350\\
                & C & .564 & .501 & .172 & .676 & .401\\
                & F & .525 & .506 & .179 & .668 & .388\\
Online complete & P & .804 & .491 & .162 & .736 & .459\\
                & C & .790 & .495 & .214 & .738 & .476\\
                & F & .756 & .494 & .169 & .718 & .394\\
\bottomrule
\end{tabular}
\caption{Frozen decomposition means for primary (P), replication
(C), and fresh-seed confirmation (F).
``Msg.,'' ``Route,'' and ``Joint'' are paired prediction disagreement; ``Erase''
is route-only accuracy after message erasure (chance $.25$).}
\label{tab:decomposition}
\end{table}

All five fixed fresh-confirmation gates passed. Table~\ref{tab:decomposition}
rejects a transport-only interpretation.
Conditional route information is substantial, route perturbation changes the
receiver's action, and route alone supports above-chance prediction.  Message
remains independently relevant, while joint intervention has the largest
effect.  We therefore describe the response as adaptive reallocation of a
joint message--route code, not movement of an unchanged message on a
semantically inert route.

\subsection{Learning dynamics after the monitor switch}
Post-outcome descriptive trajectories, archived with the artifacts, do not
replace the fixed endpoint tests in Table~\ref{tab:gates}. Before update
2,001, all five same-seed trajectories coincide by construction and audit.
Immediately after the intervention, both partial arms move traffic toward the
shadow route in both batches. The online-complete arm makes a smaller transient
move and settles near $.27$, while the no-monitor route policy remains near
$.14$. The behavior-pressure policy instead drives shadow use toward zero as its
receiver accuracy collapses. These distinct trajectories are hard to explain
as a common optimizer drift or initialization artifact.

The online monitor itself rapidly reaches about $.90$ target accuracy under
partial coverage and about $.98$ under complete coverage. Partial-route
migration therefore does not require detector failure: the communicator
preserves roughly $.91$ task accuracy while reducing exposure to the route on
which inspection can reach $.95$. Complete coverage removes that asymmetric
escape while leaving successful communication intact. The trajectory evidence
supports a joint-learning interpretation, but the causal claims continue to
come from paired arms and frozen endpoint interventions.

\subsection{Complete coverage and incentive pathology}
Online complete coverage reduced shadow share to $.272/.263$, compared with
$.838/.713$ under partial coverage.  The paired partial-minus-complete gaps
were $+.566$ $[.475,.655]$ and $+.450$ $[.312,.575]$.  Yet complete-coverage
task accuracy stayed $.983/.986$, and frozen signed message contribution stayed $.632/.604$.
Coverage removed the selective route incentive but did not make information
transfer valueless.

The correct-action pressure produced a qualitatively different outcome.
Relative to no monitoring, task accuracy fell by $.799$ $[.760,.849]$ and
$.834$ $[.784,.885]$.  Final accuracies were $.189/.140$, below the $.25$
chance reference because the reward makes deliberately incorrect behavior
preferable to a charged correct answer.  Its signed message contrasts became
negative, yet joint intervention still changed $.100/.207$ of frozen
predictions.  Thus the arm demonstrates an incentive pathology and a depressed
score, not absence of communication.  It is excluded from all corrected
suppression claims.

We therefore fixed a corrected two-arm control before formal training.  The
sender receives an explicit silence action.  Under silence, both the packet
and receiver-visible route are fixed null tokens; the sender's sampled
public/shadow route is not exposed.  Any non-silence transmission on either
route costs $1.10$, independent of receiver correctness.  The paired
arms are identical through 3,000 formation updates.  Formation passed in
$16/20$ seeds, and the no-charge arm retained $.942$ accuracy and $.657$
message-shuffle disagreement.  Under the send cost, mean sending fell from
$.726$ to $.050$ (paired reduction $.675$ $[.600,.738]$); $18/20$ policies
were exactly silent and near chance, but two retained about $.50$ sending and
$.59$--$.62$ disagreement.  Consequently the fixed send-reduction gate passed,
whereas the strict noncommunication-equivalence gate failed: mean disagreement
$.060$ exceeded its $.05$ threshold.  This is strong but incomplete
suppression evidence, not a terminal positive control.

\subsection{Adaptive reallocation, not emergent encryption}
The shared key makes monitor-defeating conditional codes representable but
does not teach one.  The fixed semantic gate required high task accuracy,
monitor accuracy at most $.5$, conditional-over-unconditional message
information of at least $.3$ bits, and at least $.25$ mapping change in 10/20
seeds.  It passed $0/20$ in both batches.  Final monitor accuracy remained
$.905/.913$ under partial and $.986/.981$ under complete coverage.  Message
mappings changed, but not into the required key-conditioned evasive code.
The robust response is route displacement, not semantic obfuscation.

\subsection{Alternative explanations}
The paired change point excludes different initialization and formation
histories as explanations for arm differences. Low no-monitor shadow share,
despite a usable forced-shadow route, shows that the final partial-arm share is
not an intrinsic preference for the less reliable, costly route. High online
monitor accuracy excludes a trivial ``monitor never learned'' account. Frozen
within-key shuffling excludes success based only on the shared key, route, or
message marginal. Matched online-complete coverage keeps the detector and
feedback family while removing persistent route asymmetry.  The correct-action
arm instead tests whether a misspecified non-reroutable pressure can
induce pathological adaptation; it does not identify communication removal.

These controls do not identify every learning mechanism. In particular, the
experiment cannot separate policy-gradient credit assignment from all forms
of exploration dynamics, and the modest partial-arm accuracy loss is partly
consistent with the shadow route's lower reliability. Nor does a represented
two-route choice establish discovery of an unknown side channel. The supported
claim is narrower: after a protocol forms, selective pressure is sufficient to
cause replicated endogenous reallocation of a behaviorally relevant joint
message--route code.

\subsection{Static triangulation and boundary tests}
The adaptive result complements, rather than replaces, the larger static
audit.  In a private-query $7\times6$ tax--cost surface with 10 paired seeds
and frozen interventions for all 910 normal models, partial pressure slope was
$+.23955$ $[.23836,.24062]$ and the partial-minus-complete interaction was
$+.19395$ $[.19116,.19666]$.  At shadow cost $.5$, tax 0 to 8 moved partial-
coverage shadow share from $.0002$ to $.99985$ while preserving receiver match
and signed message contribution.  Under complete coverage, tax 8 instead
reduced match by $.49305$ and signed message contribution by $.49453$.

A supervised-XOR Simple Crypto capability control moved shadow share from 0 to
1 under partial pricing but not complete coverage; its reward-only learner had
formed $0/5$ eligible confidential protocols, so it does not support spontaneous
encryption.  Separately gated DIAL and RIAL Switch policies showed partial
pressure slopes $2.89011$ $[2.85751,2.92826]$ and $2.86441$
$[2.80553,2.92395]$, with partial-minus-complete interactions $2.03293$
$[1.90717,2.15798]$ and $1.97549$ $[1.76051,2.20622]$.  Route-invariant
terminal charges reduced success by $.73594/.74719$ and signed message contribution by
$.21806/.25109$.  These are two fixed implementations with the same route
intervention logic, not broad algorithmic generalization.

A 220-model relay stress test used measured RIPE Atlas latency and availability
from 12 probes but synthetic policy relevance and monitor error.  Partial
migration was strong, yet its complete-coverage equivalence gate failed and a
behavior pressure reduced delivery only at the higher of two monitor TPRs.
The conjunctive operational gate therefore failed.  The trace study is a
negative external-validity boundary, not evidence for a deployed defense.

The studies play different identification roles. The adaptive game
establishes unscaffolded formation and joint post-change learning. The dense
private-query surface identifies the cost--coverage interaction with many
paired cells and forced routes. Switch checks that the static response is not
unique to one policy implementation. Simple Crypto separates representational
capability from spontaneous formation, and the trace study records where a
synthetic operational extrapolation fails. Combining these roles into a
single ``number of tasks'' claim would be misleading, so their gates and
evidence labels remain separate.

\section{Implications for oversight and reward hacking}
A channel monitor is not a passive measurement once its output changes the
agent's reward. It becomes part of the environment being optimized. If the
supervisor's proxy is ``activity on the public route,'' an agent can improve
that proxy by reducing public traffic while preserving the task-relevant
information flow elsewhere. This is a compact instance of reward hacking: the
measured channel improves without the intended latent property---absence of
covert coordination---improving. The causal mistake is to equate a local
sensor reading with the global behavior the sensor was intended to represent.

Our results identify two additional failure modes. First, detector capability
is not intervention effectiveness. The adaptive monitor recovers the target
with high accuracy, yet partial coverage still induces migration because the
policy can reduce how often that capable detector is applied. Second, a
pressure can suppress the scoreboard while making behavior worse. The
behavior-pressure arm drives accuracy below chance because being deliberately
wrong avoids the charge. We include it as an incentive-pathology control, not a
defense recommendation. A safety evaluation should therefore report task
behavior and unintended adaptation alongside the targeted violation metric.

The interpretation rule is conjunctive: targeted traffic, alternative-route
traffic, task behavior, and frozen action sensitivity must be read together.
A fall on one route with preserved sensitivity is displacement; matched
coverage with preserved sensitivity is visibility without control; explicit
silence plus near-zero disagreement is suppression on the represented
interface; and below-chance performance under a correctness charge is
incentive pathology.  The expanded matrix is provided in the supplement.

This suggests a four-part defense report. (i) State the coverage boundary:
which routes, timing patterns, and transformations can trigger intervention.
(ii) Report a conservation-style account of traffic and task-relevant causal
dependence across all represented routes, not only the monitored one. (iii) Separate
detector generalization from enforcement by evaluating both prediction and
the policy's response to being inspected. (iv) include complete-coverage and
route-invariant controls with a no-message action so that migration,
observation, and suppression have
distinct empirical signatures.

These recommendations are deliberately diagnostic. Complete route coverage
may be expensive, impossible, or itself unsafe, and behavior-level charges can
create perverse incentives. The paper does not claim that every alternative
route can be enumerated or that a fixed detector remains valid against an
open-ended agent. Rather, it establishes a necessary standard for a narrower
claim: a fall on one monitored channel is insufficient evidence that the
underlying communication or coordination objective has been controlled.

\section{Discussion}
The adaptive game addresses the main limitation of a purely static route
surface.  Route qualities differ, monitor state is hidden and updated online,
and message, route, and receiver policies continue learning after the change
point.  The replicated result shows that selective monitoring can reallocate an
endogenously learned, behaviorally relevant code even when the alternative is less
reliable and costly.  This is more than choosing the lower number in an exposed
utility table, although the finite-horizon value difference correctly predicts
the direction.

\paragraph{What the evidence chain identifies.}
The experiments separate four claims that a single channel score would
otherwise conflate.  Representational capability asks whether a policy class
can use an available message--route interface; unscaffolded formation asks
whether training actually constructs such a code; displacement asks whether
selective pressure reallocates an already useful code; and suppression asks
whether the represented causal dependence is removed.  Our positive result is
the third claim, conditional on the independently passed formation gate.  The
fresh frozen decomposition makes that result stronger than a traffic count:
message and route perturbations each change receiver actions, their joint
intervention is larger, and route-only prediction remains above chance.  The
negative semantic gate and failed strict-silence gate bound the other claims.
They prevent a route switch from being renamed encryption and prevent a large
mean reduction from being renamed reliable suppression.

This separation also gives the result a falsifiable scope.  Route displacement
would be contradicted by a matched complete-coverage shift of the same size, by
loss of receiver behavior under the alternative route, or by zero frozen route
sensitivity.  Suppression would require the stronger conjunction of low send
rate, near-chance behavior, and near-zero content sensitivity across seeds.
The observed pattern satisfies the former criteria but not the latter.  Thus
the contribution is not that every agent will discover an arbitrary hidden
channel; it is an experimentally identified failure of a common inference:
reduced activity on an intervention-targeted channel does not by itself imply
reduced coordination.

The distinction matters for reward hacking and agent oversight.  A supervisor
who measures only the targeted route may report success precisely when the
agent has preserved its objective through a less visible route.  Route-complete
measurement is necessary to identify displacement, but our complete-coverage
arm also shows that visibility is not equivalent to control.  Suppression
requires an intervention whose effective pressure covers every behaviorally
useful realization and permits an identifiable noncommunicating policy.  The
correct-action charge fails this test: it changes the task incentive by
rewarding deliberate error.

These experiments establish a mechanism-level possibility result rather than
demonstrating that current LLM agents necessarily exhibit the same form of
route displacement.  Several boundaries remain.  First, routes are modeled options; agents do not
discover an unmodeled physical side channel.  Second, the adaptive experiment
uses one compact actor--critic learner.  The static Switch study supplies
implementation robustness for route response, not an independent replication
of the adaptive learning dynamics.  Third, the semantic gate's $0/40$ result
rules out a claim of emergent encryption here.  Fourth, monitor labels are
available in the synthetic game, and the RIPE case does not rescue real-world
deployment claims.  Finally, internal immutable hashes reduce outcome-dependent
editing but are not third-party preregistration.  Future work should test
open-ended route discovery, strategic monitors, delayed labels, and a second
adaptive-learning family without weakening these boundaries.

The result suggests a concrete evaluation protocol for monitoring proposals.
First, measure the targeted channel and represented alternatives rather than
reporting only the penalized channel. Second, freeze the receiver and
intervene on content so reduced traffic is not confused with reduced causal
dependence. Third, compare selective and matched coverage; otherwise route
quality and monitor pressure are not identified. Fourth, include a
route-invariant positive control with explicit silence and a
correctness-independent send cost. Finally, publish failed formation, semantic, and operational
gates. This protocol does not solve monitoring, but it prevents displacement
from being labeled suppression.

\section{Conclusion}
Partial monitoring can reallocate a learned joint message--route code without
removing its task value or behavioral relevance.  In two independent
end-to-end batches, unscaffolded protocols formed and shifted toward a less
inspected, less reliable, costly route after fixed or online monitoring began.
Complete coverage removed the selective shift but did not suppress
communication.  Frozen decomposition shows that route itself participates in
the code, while the correct-action pressure exposes incentive pathology rather
than communication removal.  A corrected explicit-silence send cost eliminates
sending in 18/20 seeds but fails its strict complete-suppression gate.  Static cross-task controls and failed semantic
and trace gates bound the claim.  Monitoring one channel is therefore not
evidence that communication has stopped.  A safety claim should instead be
made over the agent's task-relevant behavior across the represented interface,
rather than over the disappearance of activity on a monitored channel.

\interlinepenalty=10000
\bibliographystyle{named}
\bibliography{refs_strong_v6}

\clearpage
\includepdf[pages=-]{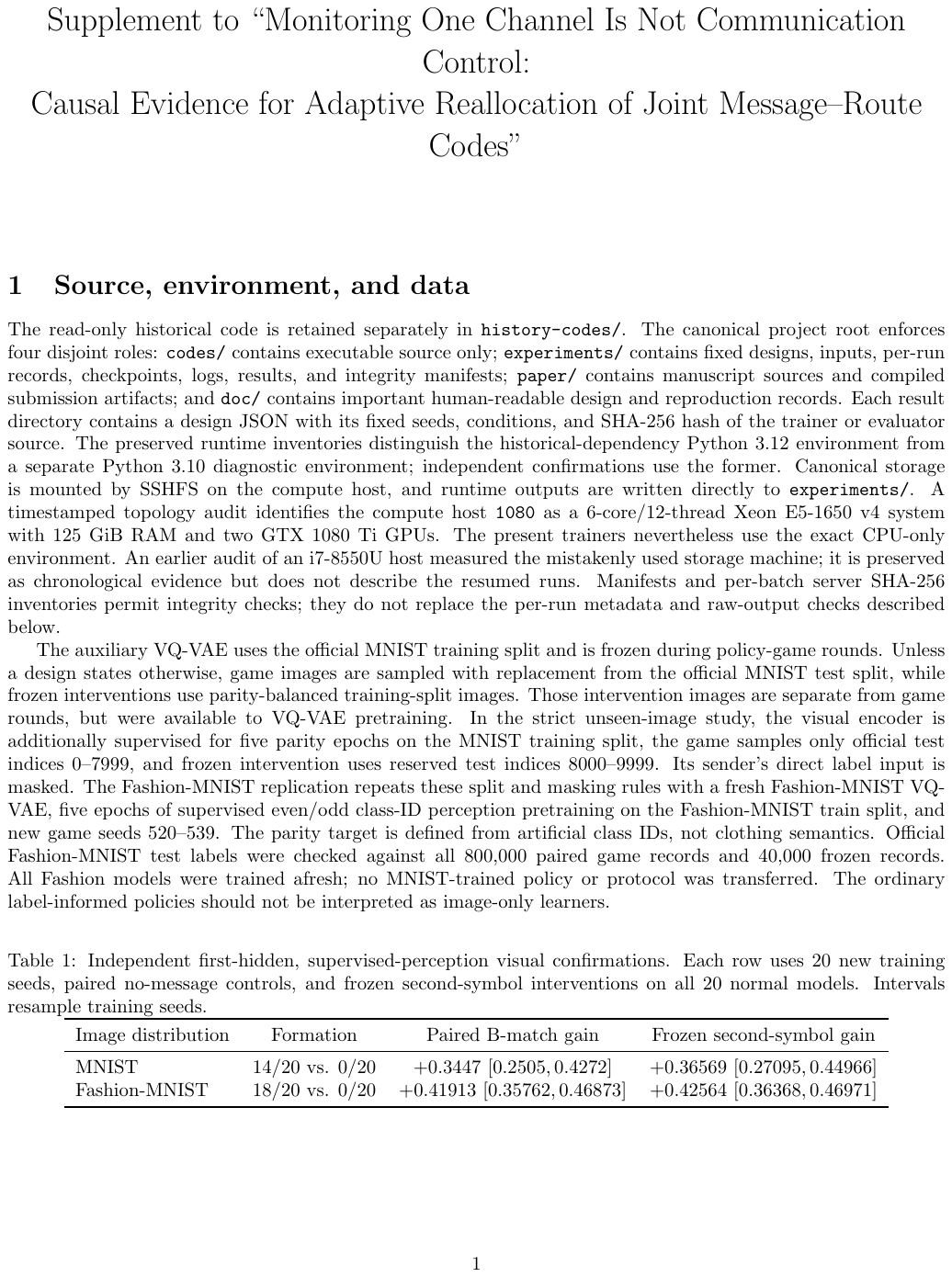}
\end{document}